\documentclass[runningheads]{llncs}
\usepackage[T1]{fontenc}
\usepackage{graphicx,verbatim}
\usepackage{booktabs}
\usepackage{multirow}
\usepackage{amsmath}
\usepackage{amssymb}
\usepackage[table,xcdraw]{xcolor}
\usepackage{url}
\newcommand{\CBF}{\mathrm{CBF}}
\newcommand{\CBV}{\mathrm{CBV}}
\newcommand{\MTT}{\mathrm{MTT}}
\newcommand{\Tmax}{T_{\max}}
\newcommand{\AIF}{\mathrm{AIF}}
\newcommand{\normone}[1]{\lVert #1 \rVert_1}

\begin{document}
\title{Evidential Perfusion Physics-Informed Neural Networks with Residual Uncertainty Quantification}

\titlerunning{EPPINN: Evidential Perfusion PINNs}
%

\author{
Junhyeok Lee\inst{1} \and
Minseo Choi\inst{2} \and
Han Jang\inst{3} \and
Young Hun Jeon\inst{2} \and
Heeseong Eum\inst{1} \and
Joon Jang\inst{4} \and
Chul-Ho Sohn\inst{2} \and
Kyu Sung Choi\inst{2,5,6}
}
\authorrunning{J. Lee et al.}
\institute{
Cancer Biology, Seoul National University College of Medicine, Seoul, Korea \and
Radiology, Seoul National University Hospital, Seoul, Korea \and
Bioengineering, Seoul National University, Seoul, Korea \and
Biomedical Sciences, Seoul National University, Seoul, Korea \and
Radiology, Seoul National University College of Medicine, Seoul, Korea \and
Healthcare AI Research Institute, Seoul National University Hospital, Seoul, Korea\\
\email{evening0619@gmail.com, ent1127@snu.ac.kr}
}
\maketitle

\begin{abstract}
Physics-informed neural networks (PINNs) have shown promise in addressing the ill-posed deconvolution problem in computed tomography perfusion (CTP) imaging for acute ischemic stroke assessment. However, existing PINN-based approaches remain deterministic and do not quantify uncertainty associated with violations of physics constraints, limiting reliability assessment. We propose Evidential Perfusion Physics-Informed Neural Networks (EPPINN), a framework that integrates evidential deep learning with physics-informed modeling to enable uncertainty-aware perfusion parameter estimation. EPPINN models arterial input, tissue concentration, and perfusion parameters using coordinate-based networks, and places a Normal--Inverse--Gamma distribution over the physics residual to characterize voxel-wise aleatoric and epistemic uncertainty in physics consistency without requiring Bayesian sampling or ensemble inference. The framework further incorporates physiologically constrained parameterization and stabilization strategies to promote robust per-case optimization. We evaluate EPPINN on digital phantom data, the ISLES 2018 benchmark, and a clinical cohort. On the evaluated datasets, EPPINN achieves lower normalized mean absolute error than classical deconvolution and PINN baselines, particularly under sparse temporal sampling and low signal-to-noise conditions, while providing conservative uncertainty estimates with high empirical coverage. On clinical data, EPPINN attains the highest voxel-level and case-level infarct-core detection sensitivity. These results suggest that evidential physics-informed learning can improve both accuracy and reliability of CTP analysis for time-critical stroke assessment. Source code is available at \url{https://github.com/jhlee0619/EPPINN}.
\end{abstract}

\section{Introduction}
Acute ischemic stroke management is time-critical, with reperfusion therapies indicated within 4.5 hours for intravenous thrombolysis and within 6 hours for mechanical thrombectomy, extending up to 24 hours in selected patients~\cite{saver2006time,albers2018thrombectomy}. Rapid differentiation between the irreversible ischemic core and the salvageable penumbra is therefore essential for triage~\cite{goyal2020challenging}. Computed Tomography Perfusion (CTP) is widely used to quantify hemodynamic parameters such as cerebral blood flow ($\CBF$), cerebral blood volume ($\CBV$), and mean transit time ($\MTT$)~\cite{alves2014reliability}. This quantification relies on deconvolution, an ill-posed inverse problem sensitive to noise and arterial input function ($\AIF$) variability~\cite{fieselmann2011deconvolution}. Traditional singular value decomposition (SVD) pipelines are efficient but unstable under low signal-to-noise ratios or sparse temporal sampling~\cite{ostergaard1996high,wu2003tracer}. Although deep learning models improve robustness by learning direct mappings from time-attenuation curves~\cite{livne2019u}, purely data-driven approaches may violate tracer-kinetic constraints and yield physiologically implausible or poorly generalizable estimates across acquisition protocols, potentially compromising time-critical treatment decisions.

Physics-informed neural networks (PINNs) incorporate tracer-kinetic equations as soft constraints and have improved numerical stability in CTP through spatio-temporal formulations that explicitly enforce the convolution model~\cite{raissi2019physics,de2023spatio,de2024accelerating}. Nevertheless, existing PINN-based approaches remain deterministic and yield only point estimates, without quantifying uncertainty arising from imperfect physics constraint satisfaction—an important limitation in acute stroke triage, where treatment decisions depend on reliable parameter confidence~\cite{albers2018thrombectomy}. This limitation is further amplified in clinical CTP, which is inherently noisy and often sparsely sampled in time, increasing residual ambiguity and destabilizing optimization. Although Bayesian PINNs and ensemble methods can provide uncertainty estimates~\cite{yang2021b,lakshminarayanan2017simple}, their computational overhead limits practicality for rapid, per-case clinical deployment.

To this end, we propose Evidential Perfusion Physics-Informed Neural Networks (EPPINN), an uncertainty-aware framework that integrates evidential learning~\cite{sensoy2018evidential,amini2020deep} with physics-informed modeling for CTP analysis. Unlike deterministic PINNs, EPPINN introduces a probabilistic formulation of physics constraint deviations to enable explicit uncertainty estimation while retaining a single optimization procedure suitable for per-case deployment. The framework further incorporates optimization stabilization strategies to ensure robust performance under noisy and sparsely sampled clinical conditions. We validate the proposed approach on synthetic and clinical datasets.

Our contributions are summarized as follows:
\begin{itemize}
\item We introduce an uncertainty-aware physics-informed framework for CTP that models tracer-kinetic constraint deviations probabilistically, yielding voxel-wise aleatoric and epistemic uncertainty estimates without additional computational overhead.
\item We propose a stable per-case optimization scheme that integrates evidential residual learning with a structured $\CBV$--$\MTT$ parameterization and progressive annealing, ensuring robust convergence in noisy clinical settings.
\item Through extensive evaluation on digital phantoms, the ISLES 2018 benchmark, and clinical data, we demonstrate that EPPINN improves perfusion parameter accuracy and infarct-core detection sensitivity compared to classical deconvolution and state-of-the-art PINN baselines.
\end{itemize}

\section{Method}
\label{sec:method}

\begin{figure*}[t]
\centering
\includegraphics[width=\textwidth]{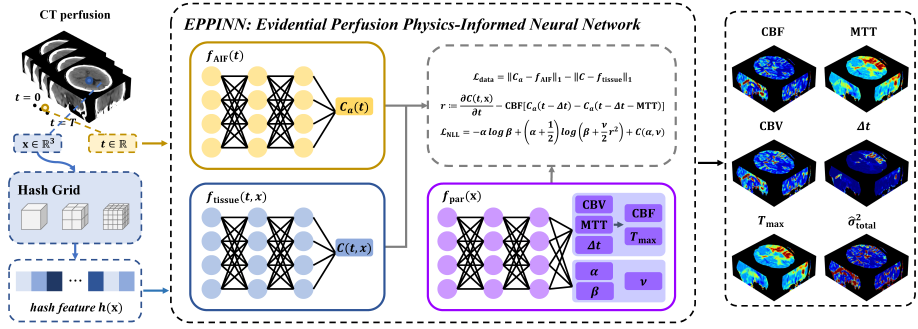}
\caption{Overview of EPPINN. Coordinate-based networks model AIF and tissue curves, and a parameter network predicts perfusion parameters and evidential residual uncertainty.}
\label{fig:overview}
\end{figure*}

\subsection{CTP Forward Model and Physics Constraints}
\label{sec:method_prelim}

Let $t\in\mathbb{R}$ denote continuous time and $\mathbf{x}\in\mathbb{R}^3$ the voxel coordinate.
In CTP, the tissue concentration curve $C(t,\mathbf{x})$ at voxel $\mathbf{x}$ is related to the AIF $C_a(t)$ through the convolution model~\cite{ostergaard1996high}
\begin{equation}
C(t,\mathbf{x})
=
\CBF(\mathbf{x}) \int_{0}^{t}
C_a(\tau-\Delta t(\mathbf{x}))\, R(t-\tau,\mathbf{x})\, d\tau,
\label{eq:ct_convolution}
\end{equation}
where $\CBF(\mathbf{x})$ denotes cerebral blood flow, $\Delta t(\mathbf{x})$ the transit delay, and $R(\cdot,\mathbf{x})$ the residue function, so that the impulse response function (IRF) is given by $\mathrm{IRF}(t,\mathbf{x}) = \CBF(\mathbf{x}) R(t,\mathbf{x})$.
Following prior spatio-temporal PINN-based CTP formulations~\cite{de2023spatio,de2024accelerating}, we adopt a box residue function, i.e., $R(s,\mathbf{x})=1$ for $0\le s\le \MTT(\mathbf{x})$ and $0$ otherwise. Substituting this into Eq.~(\ref{eq:ct_convolution}) and applying the Leibniz rule yields the endpoint-difference constraint and physics residual:
\begin{align}
\frac{\partial C(t,\mathbf{x})}{\partial t}
&=
\CBF(\mathbf{x})\Big[
C_a\!\big(t\!-\!\Delta t(\mathbf{x})\big)
\!-\!
C_a\!\big(t\!-\!\Delta t(\mathbf{x})\!-\!\MTT(\mathbf{x})\big)
\Big],
\label{eq:physics_constraint} \\
r(t,\mathbf{x})
&:=
\frac{\partial C(t,\mathbf{x})}{\partial t}
-
\CBF(\mathbf{x})\Big[
C_a\!\big(t\!-\!\Delta t(\mathbf{x})\big)
\!-\!
C_a\!\big(t\!-\!\Delta t(\mathbf{x})\!-\!\MTT(\mathbf{x})\big)
\Big].
\label{eq:residual}
\end{align}

Under the governing constraint, the residual $r(t,\mathbf{x})$ should vanish. 
To enforce this constraint within a neural approximation framework, 
let $f_{\mathrm{AIF}}(t)$ and $f_{\mathrm{tissue}}(t,\mathbf{x})$ denote neural approximations of $C_a(t)$ and $C(t,\mathbf{x})$, respectively (see Sec.~\ref{sec:method_model}). 
In practice, we evaluate $r(t,\mathbf{x})$ by substituting $C$ with $f_{\mathrm{tissue}}$ and computing the temporal derivative via automatic differentiation.

In addition to the physics residual, we define an $L_1$ data-fidelity loss on the arterial input function and tissue concentration curves, evaluated at the acquired time points and sampled voxel locations:
\begin{equation}
\mathcal{L}_{\mathrm{data}}
=
\normone{C_a - f_{\mathrm{AIF}}} + \normone{C - f_{\mathrm{tissue}}}.
\label{eq:data_loss}
\end{equation}

\subsection{Architecture and Evidential Formulation}
\label{sec:method_model}

EPPINN consists of three coordinate-based networks (Fig.~\ref{fig:overview}): the AIF and tissue networks $f_{\mathrm{AIF}}$ and $f_{\mathrm{tissue}}$, and a parameter network $f_{\mathrm{par}}$ that predicts $\CBV$, $\MTT$, and $\Delta t$. The flow parameter $\CBF$ is derived from $(\CBV, \MTT)$ following the $\CBV$--$\MTT$ parameterization (Sec.~\ref{sec:method_stability}). Spatial coordinates $\mathbf{x}$ are encoded by a multi-resolution hash grid~\cite{muller2022instant} $h(\mathbf{x})$. The tissue network takes $[t, h(\mathbf{x})]$ as input, and the parameter network takes $h(\mathbf{x})$ as input. For brevity, we write $f_{\mathrm{tissue}}(t,\mathbf{x})$ and $f_{\mathrm{par}}(\mathbf{x})$.

To enable uncertainty estimation, $f_{\mathrm{par}}$ additionally outputs evidential parameters that parameterize a Normal--Inverse--Gamma (NIG) distribution over the physics residual $r(t,\mathbf{x})$. 
The NIG formulation provides computationally efficient, closed-form uncertainty decomposition under the prior $\sigma^2\!\sim\!\mathrm{Inv\Gamma}(\alpha,\beta)$ and $\mu\!\sim\!\mathcal{N}(0,\sigma^2/\nu)$~\cite{amini2020deep}.
Specifically, the network predicts $(\tilde{\alpha},\tilde{\beta},\tilde{\nu})$, which are transformed to valid domains by:
$\alpha = 1+\mathrm{softplus}(\tilde{\alpha})+\epsilon_{\text{edl}}$,
$\beta=\mathrm{softplus}(\tilde{\beta})+\epsilon_{\text{edl}}$,
$\nu=\mathrm{softplus}(\tilde{\nu})+\epsilon_{\text{edl}}$.

\textbf{Evidential residual learning and uncertainty.}
We treat the physics residual as an evidential regression target with $y_r:=0$ under this NIG prior~\cite{amini2020deep}. Since the governing constraint requires $r(t,\mathbf{x})=0$, fixing the predictive mean ($\gamma=0$) enforces physics consistency while allowing $(\alpha,\beta,\nu)$ to encode uncertainty in constraint satisfaction. 
We optimize the standard NIG negative log-likelihood with an anti-degeneracy regularizer:
\begin{equation}
\mathcal{L}_{\mathrm{NLL}}
=
-\alpha\log\beta
+
\left(\alpha+\tfrac{1}{2}\right)
\log\!\left(\beta+\tfrac{\nu}{2}r^2\right)
+
C(\alpha,\nu),
\end{equation}
where $C(\alpha,\nu)$ collects terms independent of $r$, and
\begin{equation}
\mathcal{L}_{\mathrm{reg}} = |r|\,(2\nu+\alpha).
\end{equation}
The NIG prior promotes small residuals while quantifying remaining discrepancies as aleatoric and epistemic components. Voxel-wise uncertainty maps are obtained via the closed-form decomposition~\cite{amini2020deep}:
\[
\hat{\sigma}^2_{\mathrm{ale}}=\frac{\beta}{\alpha-1}, \quad
\hat{\sigma}^2_{\mathrm{epi}}=\frac{\hat{\sigma}^2_{\mathrm{ale}}}{\nu}, \quad
\hat{\sigma}^2_{\mathrm{total}}=\hat{\sigma}^2_{\mathrm{ale}}+\hat{\sigma}^2_{\mathrm{epi}},
\]
providing physics-consistency uncertainty in a single forward pass.

\textbf{Objective and annealing.}
The overall objective combines data fidelity (Eq.~\ref{eq:data_loss}), residual minimization, and evidential regularization:
\begin{equation}
\mathcal{L}
=
\lambda_{\mathrm{data}}\,\mathcal{L}_{\mathrm{data}}
+
\lambda_{\mathrm{res}}
\left(
\normone{r}
+
\omega(i)\,\lambda_{\mathrm{EDL}}
\big[
\mathcal{L}_{\mathrm{NLL}} + \lambda_{\mathrm{reg}}\mathcal{L}_{\mathrm{reg}}
\big]
\right),
\label{eq:total_loss}
\end{equation}
where $\omega(i)\in[0,1]$ is a progressive annealing schedule that gradually increases the influence of the evidential terms, enabling initial curve fitting before enforcing physics consistency and thereby improving optimization stability.

\subsection{Parameterization and Optimization Stability}
\label{sec:method_stability}
To improve numerical stability and identifiability in the ill-posed inverse problem, we incorporate structured parameterization and training strategies.
\textbf{(1) CBV--MTT parameterization.}
The perfusion parameters satisfy the Central Volume Principle~\cite{ostergaard1996high} $\CBV(\mathbf{x}) = \CBF(\mathbf{x}) \cdot \MTT(\mathbf{x})$. 
We therefore predict $\CBV$ and $\MTT$ as primary outputs and derive $\CBF(\mathbf{x})=\CBV(\mathbf{x})/(\MTT(\mathbf{x})+\epsilon_{\text{cbf}})$ by construction. This enforces physiological consistency and reduces ambiguity between flow and transit time parameters. The time-to-maximum ($\Tmax$) is derived as $\Tmax(\mathbf{x}) = \Delta t(\mathbf{x}) + \frac{1}{2}\MTT(\mathbf{x})$.
\textbf{(2) FOV-aware adaptive encoding.}
Spatial coordinates are normalized by the physical field-of-view prior to hash encoding. This maintains consistent effective frequency representation across voxel resolutions and scanning protocols, mitigating scale-dependent instability during optimization.
\textbf{(3) AIF pre-training.}
The AIF network $f_{\mathrm{AIF}}$ is first fitted to the global arterial signal. This reduces gradient variance during early training and stabilizes subsequent joint optimization.
\textbf{(4) Physics-informed initialization.}
The parameter network $f_{\mathrm{par}}$ is initialized to physiologically plausible values (e.g., $\CBF \approx 20$\,ml/100g/min), anchoring optimization in a realistic regime and reducing convergence to implausible local minima.
Together, these strategies improve robustness under sparse temporal sampling and heterogeneous acquisition protocols.

\section{Experiments}
\label{sec:experiments}

\subsection{Datasets}
\label{sec:datasets}
We evaluated our method on three datasets:
(i) a digital brain perfusion phantom dataset~\cite{aichert2013realistic,riordan2011validation} with known ground-truth perfusion and timing parameters across controlled noise and temporal sampling regimes, (ii) the ISLES 2018 benchmark~\cite{maier2017isles} (n=93) with 4D CTP and diffusion-weighted imaging (DWI) core masks (ROI evaluation), and (iii) a retrospective clinical cohort (n=42) with 4D CTP, DWI core masks, and AIF signals. AIF ground truth is from manual ROIs (clinical), RAPID-derived curves~\cite{de2023spatio} (ISLES), and the analytic curve (phantom); cohorts are partitioned at the patient level. The clinical scans were acquired with in-plane resolution ~0.4\,mm, slice thickness 10\,mm, temporal sampling interval ~2.5\,s (range 1.0--2.8\,s), and ~36 time points.

\begin{figure*}[t]
\centering
\includegraphics[width=0.9\textwidth]{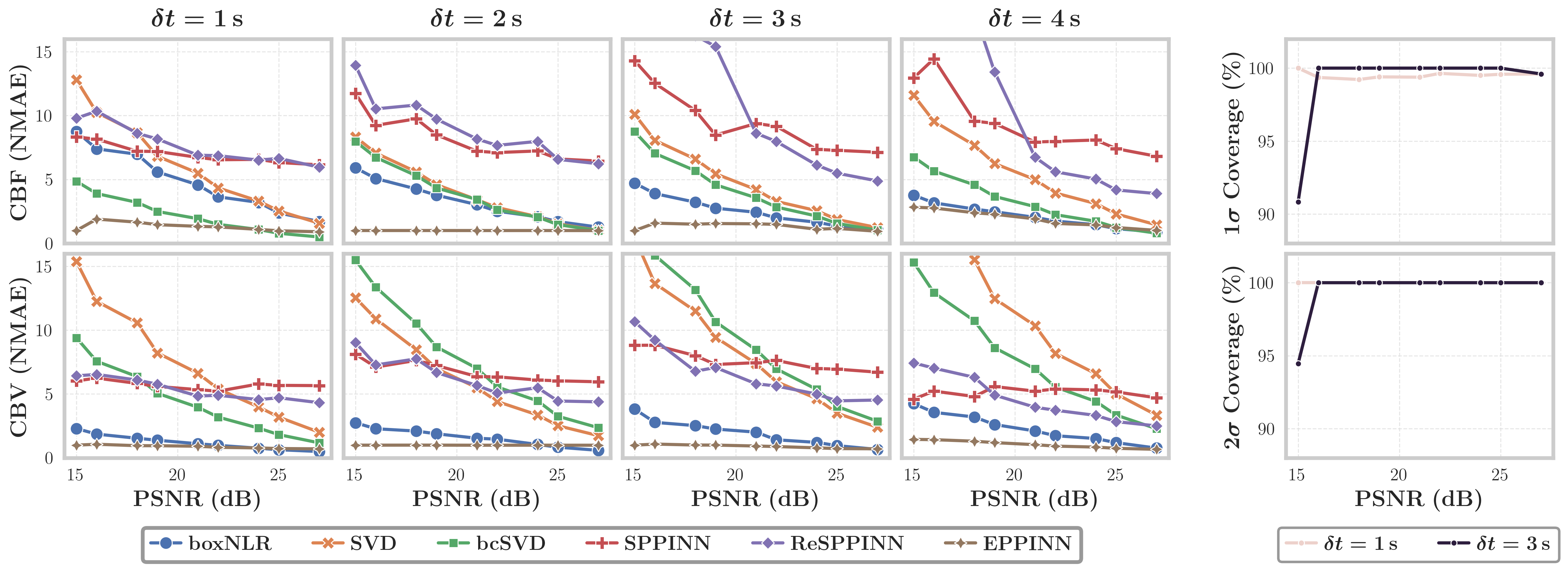}
\caption{Digital phantom results under controlled noise and temporal sampling. Left: NMAE for $\CBF$/$\CBV$ across PSNR and $\delta t$. Right: empirical coverage of uncertainty intervals. Spearman $\rho(\sigma_r,|r|)=0.66$ with $\pm1\sigma$/$\pm2\sigma$ coverage $87.0\%$/$95.6\%$ (conservative, monotone).}
\label{fig:nmae_psnr}
\end{figure*}

\subsection{Baselines and Implementation Details}
\label{sec:baselines_impl}
We compare against standard SVD~\cite{ostergaard1996high}, block-circulant SVD (bcSVD)~\cite{wu2003tracer}, box-shaped non-linear regression (boxNLR)~\cite{bennink2016fast}, SPPINN~\cite{de2023spatio}, and ReSPPINN~\cite{de2024accelerating}. 
All networks use three hidden layers with sinusoidal representation networks (SIREN; $\omega_0=15$)~\cite{sitzmann2020implicit} and multi-resolution hash encoding, with hidden dimensions 128 for tissue, 64 for parameters, and 16 for AIF. 
Optimization uses Adam with a learning rate of $10^{-3}$ with a OneCycle learning-rate schedule, 25k spatiotemporal samples per iteration, and 5k iterations per case. The AIF network is pre-trained for 5k iterations using only the $L_1$ data term (Eq.~\ref{eq:data_loss}), followed by joint optimization under the full objective (Eq.~\ref{eq:total_loss}). 
Loss weights are $\lambda_{\text{data}}=\lambda_{\text{res}}=1$, $\lambda_{\text{EDL}}=0.5$, and $\lambda_{\text{reg}}=10^{-3}$ with linear annealing, and $\epsilon_{\text{cbf}}=\epsilon_{\text{edl}}=10^{-3}$. Experiments were conducted on a single NVIDIA RTX 3090 graphics processing unit (GPU) using tiny-cuda-nn~\cite{tiny-cuda-nn}, requiring ~50\,s per case (AIF 5--6\,s, joint optimization 42\,s). Hyperparameters were fixed on the phantom and reused on ISLES and clinical as held-out tests, with learnable baselines sharing an identical budget.

\subsection{Digital Phantom Experiments}
\label{sec:synthetic}
Digital phantom experiments enable controlled evaluation under varying noise levels (PSNR 18--27\,dB) and temporal sampling intervals ($\delta t \in {1,2,3,4},\mathrm{s}$).
Accuracy is quantified using normalized mean absolute error (NMAE) across PSNR and $\delta t$ conditions, and uncertainty calibration is assessed via empirical $\pm1\sigma$ and $\pm2\sigma$ coverage (68\%/95\%).
Figure~\ref{fig:nmae_psnr} presents NMAE trends over PSNR and temporal resolutions, together with coverage results for two representative settings ($\delta t=1\,\mathrm{s}$ and $\delta t=3\,\mathrm{s}$).
EPPINN consistently achieves lower NMAE than competing methods across noise and sampling regimes, with more pronounced gains under sparse temporal sampling ($\delta t=3$--$4\,\mathrm{s}$) and low-SNR conditions.
Empirical coverage generally exceeds nominal levels, particularly outside the lowest PSNR range, indicating conservative uncertainty estimates.

\subsection{Clinical Experiments}
\label{sec:clinical}
We report voxel- and case-level detection sensitivity at a $\CBF$ threshold of 25\,ml/100g/min~\cite{murphy2007serial}; case-level sensitivity emphasizes reduction of missed cases, critical for small or heterogeneous cores~\cite{taha2015metrics}. EPPINN attains the highest voxel- and case-level sensitivity across both datasets in Table~\ref{tab:core_detection}, demonstrating superior performance on the ISLES benchmark and correctly identifying 41 of 42 clinical cases. At this threshold, EPPINN's voxel-level specificity/precision are 0.79/0.04 (ISLES) and 0.54/0.18 (clinical), in a sensitivity-prioritized regime.

\begin{figure*}[t]
\centering
\includegraphics[width=0.87\textwidth]{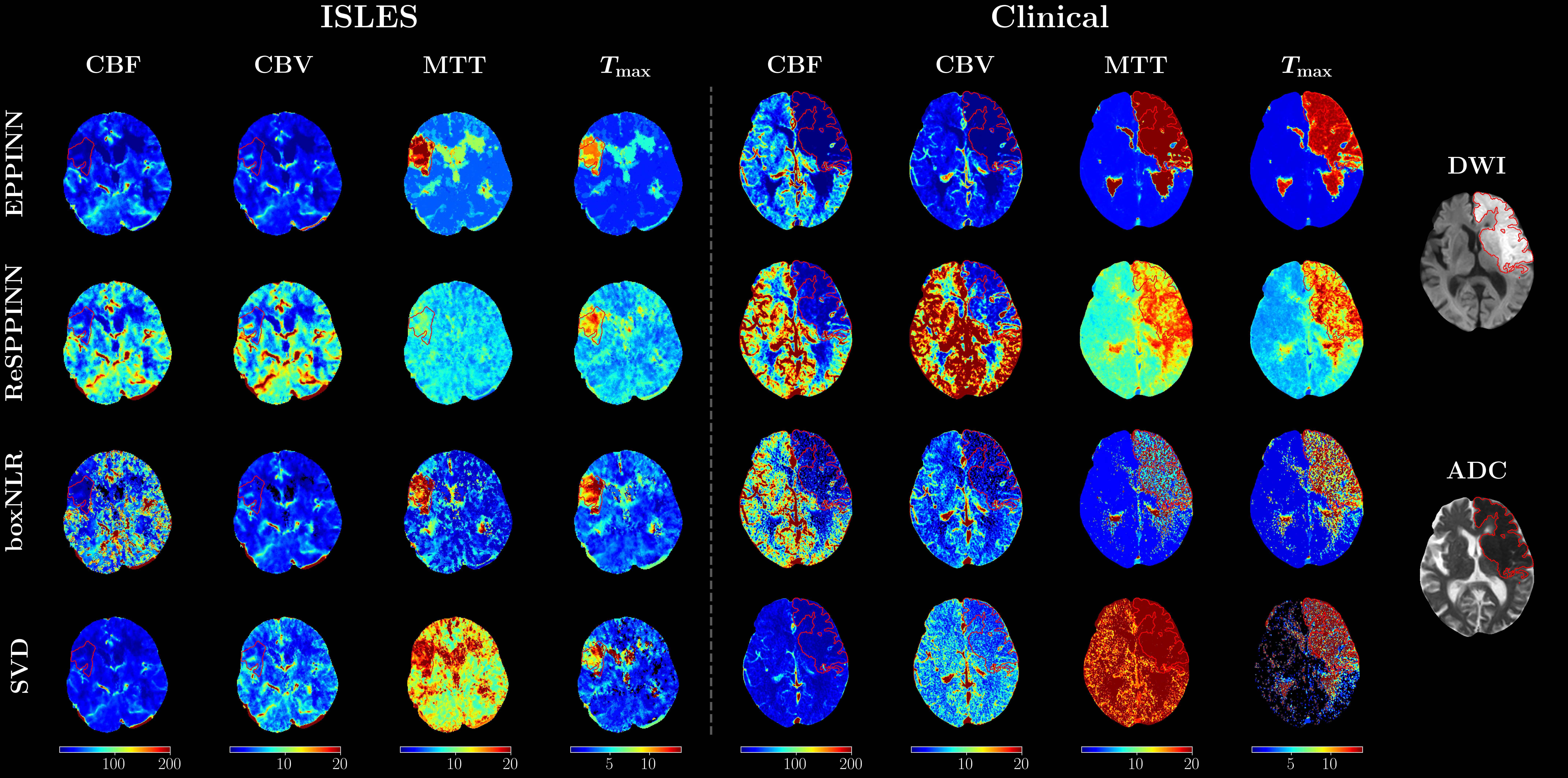}
\caption{Qualitative perfusion maps on ISLES (left) and a clinical case (right): EPPINN, ReSPPINN, boxNLR, and SVD ($\CBF$/$\CBV$/$\MTT$/$\Tmax$).}
\label{fig:qualitative}
\end{figure*}

Figure~\ref{fig:qualitative} shows qualitative comparisons. EPPINN produces coherent perfusion maps on ISLES with less speckle than ReSPPINN/boxNLR; clinical maps are spatially concordant with DWI-defined cores, while baselines show noise or over-diffusion. These qualitative observations are consistent with the improved detection sensitivity reported in Table~\ref{tab:core_detection}.

\begin{table}[t]
\centering
\caption{Infarct core detection performance on ISLES and a clinical cohort. Det. denotes a case with non-empty overlap between the predicted $\CBF\!<\!25$\,ml/100\,g/min core and the DWI core.}
\label{tab:core_detection}
\setlength{\tabcolsep}{2pt}
\setlength{\aboverulesep}{0pt}
\setlength{\belowrulesep}{2pt}
\renewcommand{\arraystretch}{1.0}
\fontsize{8}{9.5}\selectfont
\begin{tabular}{l cc c | cc c}
\toprule
\multirow{2}{*}{Method} & \multicolumn{3}{c|}{ISLES (n=93)} & \multicolumn{3}{c}{Clinical (n=42)} \\
\cmidrule(lr){2-4} \cmidrule(lr){5-7}
    & \multicolumn{2}{c}{Case-level} & Voxel-level 
    & \multicolumn{2}{c}{Case-level} & Voxel-level \\
\cmidrule(lr){2-3} \cmidrule(lr){4-4} \cmidrule(lr){5-6} \cmidrule(lr){7-7}
    & Det. / Total & Sens. $\uparrow$ & Sens. $\uparrow$
    & Det. / Total & Sens. $\uparrow$ & Sens. $\uparrow$ \\
\midrule
SVD      
    & 32 / 93 & 0.344 & 0.195
    & 12 / 42 & 0.286 & 0.125 \\
bcSVD    
    & 36 / 93 & 0.387 & 0.262
    & 27 / 42 & 0.643 & 0.395 \\
boxNLR   
    & 38 / 93 & 0.409 & 0.243
    & 38 / 42 & 0.905 & 0.406 \\
SPPINN   
    & 20 / 93 & 0.215 & 0.066
    & 29 / 42 & 0.690 & 0.304 \\
ReSPPINN 
    & 16 / 93 & 0.172 & 0.061
    & 28 / 42 & 0.667 & 0.304 \\
\midrule
\textbf{EPPINN}
    & \textbf{43 / 93} & \textbf{0.462} & \textbf{0.323}
    & \textbf{41 / 42} & \textbf{0.976} & \textbf{0.742} \\
\bottomrule
\end{tabular}
\end{table}

\begin{table}[t]
\centering
\caption{Ablation on digital phantom data ($\CBF$/$\MTT$ NMAE across ROIs). ROIs: GMR (gray matter reduced), GMSR (gray matter severely reduced), WMR (white matter reduced), WMSR (white matter severely reduced). $\dagger$: unstable training excluded from ranking.}
\label{tab:ablation_flags_cbf_mtt}
\setlength{\tabcolsep}{2pt}
\setlength{\aboverulesep}{0pt}
\setlength{\belowrulesep}{2pt}
\renewcommand{\arraystretch}{1.0}
\fontsize{8}{9.5}\selectfont
\begin{tabular*}{\textwidth}{@{\extracolsep{\fill}}lccccc|ccccc}
\toprule
    & \multicolumn{5}{c|}{$\CBF$} & \multicolumn{5}{c}{$\MTT$} \\
\cmidrule(lr){2-6}\cmidrule(lr){7-11}
Method / Strategy
& GMR & GMSR & WMR & WMSR & Avg
& GMR & GMSR & WMR & WMSR & Avg \\
\midrule
Full (proposed)
& \underline{0.66} & \underline{1.73} & \underline{1.27} & \textbf{2.76} & \underline{1.61}
& \textbf{0.59} & \textbf{0.41} & \textbf{0.64} & \textbf{0.51} & \textbf{0.54} \\

w/o Adaptive-Hash
& 0.94 & 1.83 & 1.74 & \underline{2.67} & 1.80
& 0.73 & 1.35 & 0.64 & 1.06 & 0.94 \\

w/o AIF-Pre.$^\dagger$
& 5.49 & 9.33 & 7.89 & 12.90 & 8.90
& 0.27 & 0.16 & 0.35 & 0.10 & 0.22 \\

w/o Annealing$^\dagger$
& 7.51 & 5.49 & 4.86 & 3.88 & 5.44
& 4.91 & 6.99 & 3.93 & 5.52 & 5.34 \\

w/o CBV-Param.
& 0.82 & 2.20 & 1.79 & 3.14 & 1.99
& \underline{0.71} & \underline{0.79} & \underline{0.82} & \underline{0.87} & \underline{0.80} \\

w/o Phys-init.
& \textbf{0.38} & \textbf{1.65} & \textbf{1.00} & 3.30 & \textbf{1.59}
& 0.86 & 2.06 & 0.98 & 1.75 & 1.41 \\

\bottomrule
\end{tabular*}
\end{table}

\begin{figure*}[t]
\centering
\includegraphics[width=0.9\textwidth]{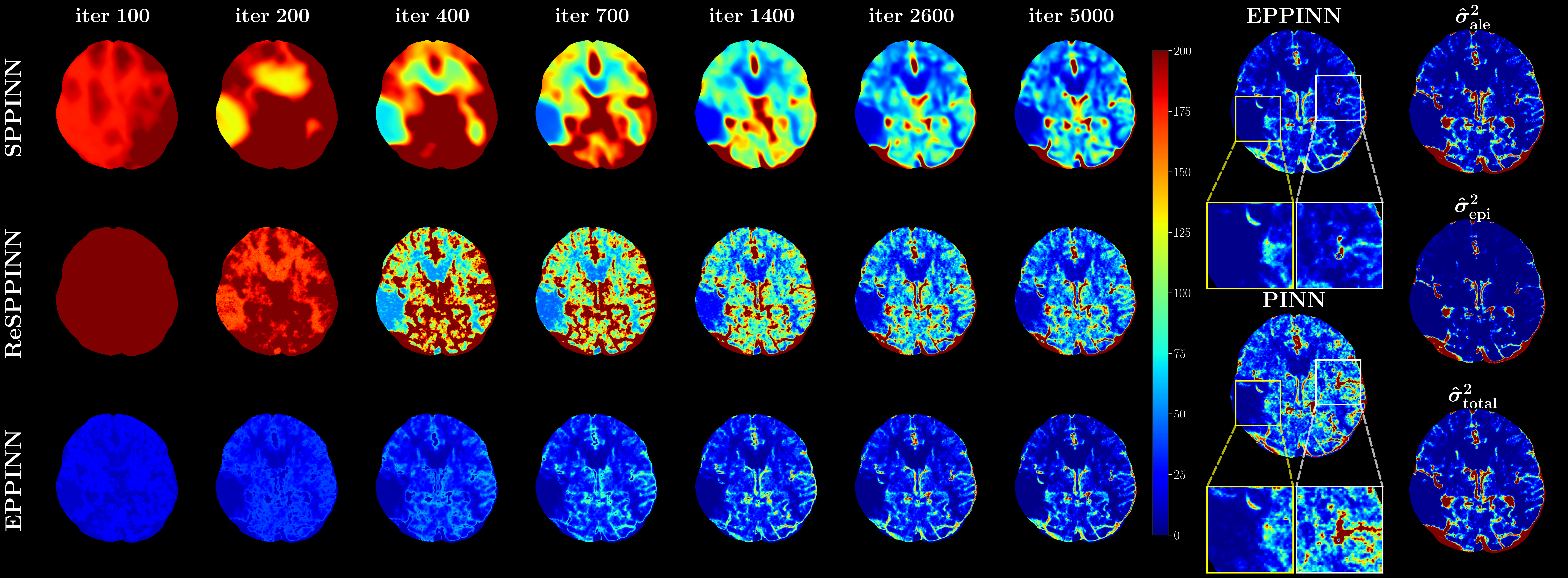}
\caption{Training dynamics and residual uncertainty. Left: iterative $\CBF$ evolution across methods. Right: final $\CBF$ comparison between EPPINN and PINN (EPPINN without evidential loss), with aleatoric, epistemic, and total residual uncertainty maps. $\hat{\sigma}_{\mathrm{ale}}$ dominates under low SNR, while $\hat{\sigma}_{\mathrm{epi}}$ is elevated near vessel/tissue boundaries.}
\label{fig:cbf_dynamics}
\end{figure*}

\subsection{Ablation Studies}
\label{sec:ablation}
Figure~\ref{fig:cbf_dynamics} illustrates training dynamics and residual uncertainty. The left panel shows that EPPINN exhibits smoother and more stable $\CBF$ convergence with reduced high-frequency fluctuations compared with the PINN baselines. The right panel compares EPPINN with its variant without the evidential loss, which produces noisier and less coherent maps; the aleatoric, epistemic, and total uncertainty localize elevated uncertainty near vascular structures and tissue boundaries.

We ablate engineering and evidential components on digital phantom data at PSNR 18 and $\delta t=1$. Performance is evaluated using NMAE normalized by the ROI mean (Table~\ref{tab:ablation_flags_cbf_mtt}).
Across ROIs, the full model achieves the best $\MTT$ performance among stable configurations, while removing $\CBV$-based parameterization consistently yields the second-best $\MTT$ results.
For $\CBF$, removing PI initialization improves performance in several regions, whereas the full model remains second-best on average with fewer convergence failures across cohorts, highlighting a stability/accuracy trade-off.

\section{Conclusion}
We presented EPPINN, an uncertainty-aware CT perfusion framework integrating evidential deep learning with physics-informed neural networks. Through evidential residual modeling with physiologically constrained parameterization, EPPINN provides voxel-wise aleatoric and epistemic confidence estimates with stable per-case optimization. Across digital phantom and clinical datasets, EPPINN achieves competitive or improved performance relative to classical deconvolution and recent PINN baselines, indicating enhanced accuracy and reliability for CT perfusion analysis in acute stroke assessment. Bayesian/ensemble UQ baselines and multicenter validation remain future work.

\begin{credits}
\subsubsection{\ackname}
This work was supported by the National Research Foundation of Korea (NRF) grant funded by the Korea government (MSIT) (No. RS-2026-25479661) (K.S.C.); the SNUH Research Fund (No. 04-2024-0600; No. 04-2025-2060) (K.S.C.); and the Korea Health Technology R\&D Project through the Korea Health Industry Development Institute (KHIDI) grant funded by the Ministry of Health\&Welfare (No. RS-2024-00439549) (K.S.C.).

\subsubsection{\discintname}
The authors have no competing interests to declare that are relevant to the content of this article.
\end{credits}


\bibliographystyle{splncs04}
\bibliography{ref}

\end{document}